\pdfoutput=1

\documentclass[11pt]{article}

\usepackage[]{emnlp2021}

\usepackage{times}
\usepackage{latexsym}

\usepackage[T1]{fontenc}

\usepackage[utf8]{inputenc}

\usepackage{microtype}
\usepackage{amsfonts,amssymb} 

\usepackage{latexsym}

\usepackage{graphicx}
\usepackage{CJKutf8}
\usepackage{multirow}
\usepackage{array}
\usepackage{tabularx}
\usepackage{makecell}
\usepackage{amsmath} 
\usepackage{amssymb}
\usepackage{tablefootnote}
\usepackage{subcaption}
\usepackage{diagbox}
\usepackage{algorithm}
\usepackage[noend]{algpseudocode}
\usepackage{booktabs}
\usepackage{xcolor}
\usepackage{color, soul} 
\usepackage{colortbl}
\usepackage{makecell}
\usepackage{amsmath}

\usepackage{microtype}
\usepackage{url}
\usepackage{booktabs}  
\usepackage{threeparttable}  
\usepackage{pgfplots}
\usepackage{arydshln}
\usepackage{float}
  
\title{Seeking Common but Distinguishing Difference,\\ A Joint Aspect-based Sentiment Analysis Model}

\author{Hongjiang Jing$^{1,2,3,\dag}$, Zuchao Li$^{1,2,3,\dag}$, Hai Zhao$^{1,2,3}$\thanks{$\ $  Corresponding author.},  and Shu Jiang$^{1,2,3}$\\

$^{1}$Department of Computer Science and Engineering, Shanghai Jiao Tong University \\
	$^{2}$Key Laboratory of Shanghai Education Commission for Intelligent Interaction \\ and Cognitive Engineering, Shanghai Jiao Tong University, Shanghai, China\\
	$^{3}$MoE Key Lab of Artificial Intelligence, AI Institute, Shanghai Jiao Tong University \\
  {\tt \small \{jinghj,charlee,jshmjs45\}@sjtu.edu.cn, zhaohai@cs.sjtu.edu.cn}}

\begin{document}

\maketitle

\begin{abstract}

Aspect-based sentiment analysis (ABSA) task consists of three typical subtasks: aspect term extraction, opinion term extraction, and sentiment polarity classification. These three subtasks are usually performed jointly to save resources and reduce the error propagation in the pipeline.
However, most of the existing joint models only focus on the benefits of encoder sharing between subtasks but ignore the difference. Therefore, we propose a joint ABSA model, which not only enjoys the benefits of encoder sharing but also focuses on the difference to improve the effectiveness of the model. In detail, we introduce a dual-encoder design, in which a pair encoder especially focuses on candidate aspect-opinion pair classification, and the original encoder keeps attention on sequence labeling.
Empirical results show that our proposed model shows robustness and signiﬁcantly outperforms the previous state-of-the-art on four benchmark datasets.

\end{abstract}

\section{Introduction}
\begin{figure}
    \centering
    \includegraphics[width=\linewidth]{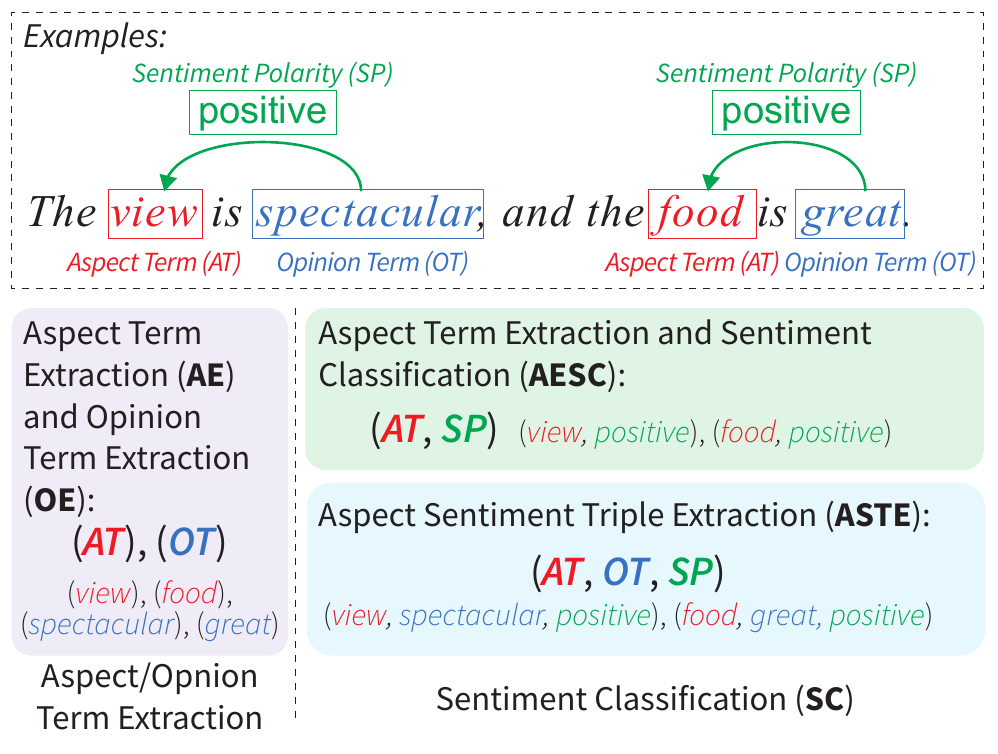}
    \caption{The subtasks in our proposed model.}
    \label{fig:task_def}
\end{figure}

Sentiment analysis is a task that aims to retrieve the sentiment polarity based on three levels of granularities: document level, sentence level, and entity and aspect level \cite{DBLP:series/synthesis/2012Liu}, which is under the urgent demands of several society scenarios \cite{DBLP:conf/cits/PreethiKOSY17,DBLP:journals/ieee-rita/CobosJB19,islam2017leveraging,novielli2018benchmark}. 
Recently, the aspect-based sentiment analysis (ABSA) task \cite{pontiki-etal-2014-semeval}, focusing on excavating the specific aspect from an annotated review, has aroused much attention from researchers, in which this paper mainly concerns the aspect/opinion term extraction and sentiment classification task.
The latest benchmark proposed by \citet{DBLP:conf/aaai/PengXBHLS20} formulates the relevant information into a triplet: target aspect object, opinion clue, and sentiment polarity orientation. 
Thus, the concerned aspect term extraction becomes a task of Aspect Sentiment Triplet Extraction (ASTE). 
Similarly, the relevant information is formulated into a pair with aspect term and sentiment polarity, and the task is defined as Aspect Term Extraction and Sentiment Classification (AESC).
Figure \ref{fig:task_def} shows an example of ASTE and AESC.

Two early methods handle the triplet extraction task efficiently \cite{zhang-etal-2020-multi-task,huang2021first}. 
Both are typically composed of a sequence representation layer to predict the aspect/opinion term mentions and a classification layer to infer the sentiment polarity of the predicted mention pair of the last layer. 
However, as is often the case, such model design may easily result in that the errors of the upper prediction layer would hurt the accuracy of the lower layer during the training procedure.

To tackle the error cascading phenomenon on the pipeline model, a growing trend of jointly modeling these subtasks in one shot appears. 
\citet{xu-etal-2020-position} proposed a joint model using a sequence tagging method, based on the bidirectional Long Short-Term Memory (LSTM) and Conditional Random Fields (CRF). 
However, they found that if a tagged mention belongs to more than one triplet, this method will be ineffective. 
\citet{zhang-etal-2020-multi-task} proposed a multi-task learning approach with the aid of dependency parsing on tail word pair of corresponding aspect-opinion pair. 
However, this non-strict dependency parsing may miss capturing structural information of term span. Meanwhile, the many-target to one-opinion issue is not effectively handled.

The promising results achieved by machine reading comprehension (MRC) frameworks on solving many other NLP tasks \cite{li-etal-2020-unified,li-etal-2019-entity} also inspires the ASTE task. 
\citet{mao2021joint} and \citet{chen2021bidirectional} attempted to design question-answer pair in terms of MRC to formulate the triplet extraction. 
Nevertheless, both need to make the converted question correspond one-to-one to the designed question manually, hence increasing computation complexity.

Among these joint models, \citet{wu-etal-2020-grid} transformed the sequence representation into the two-dimension space and argued that the word-pair under at least one assumption could represent the aspect-opinion pair as input of different encoders. 
Although this model indicated significant improvement, it treated the word-pair without taking span boundary of aspect term and opinion term into consideration and incorporated nonexistent pre-defined aspect-opinion pairs. 

Considering the problems mentioned above, we propose a dual-encoder model based on a pre-trained language model by jointly fine-tuning multiple encoders on the ABSA task.
Similar to prior work, our framework uses a shared sequence encoder to represent the aspect terms and opinion terms in the same embedding space. 
Moreover, we introduce a pair encoder to represent the aspect-opinion pair on the span level.
Thus, our dual-encoder model could learn from the ABSA subtasks individually and benefit from each other in an end-to-end manner.

Experiments on benchmark datasets show that our model significantly outperforms previous approaches at the aspect level. We also conduct a series of experiments to analyze the gain of additional representation from the proposed dual-encoder structure. 

The contributions of our work are as follows:

\indent$\bullet$ We propose a jointly optimized dual-encoder model for ABSA to boost the performance of ABSA tasks. 
   
\indent$\bullet$ We apply an attention mechanism to allow information transfer between words to promote the model to know the word pairs before inference. 
    
\indent$\bullet$ We achieve state-of-the-art performance on benchmark datasets at the time of submission.

\begin{figure*}
    \centering
    \includegraphics[width=\linewidth]{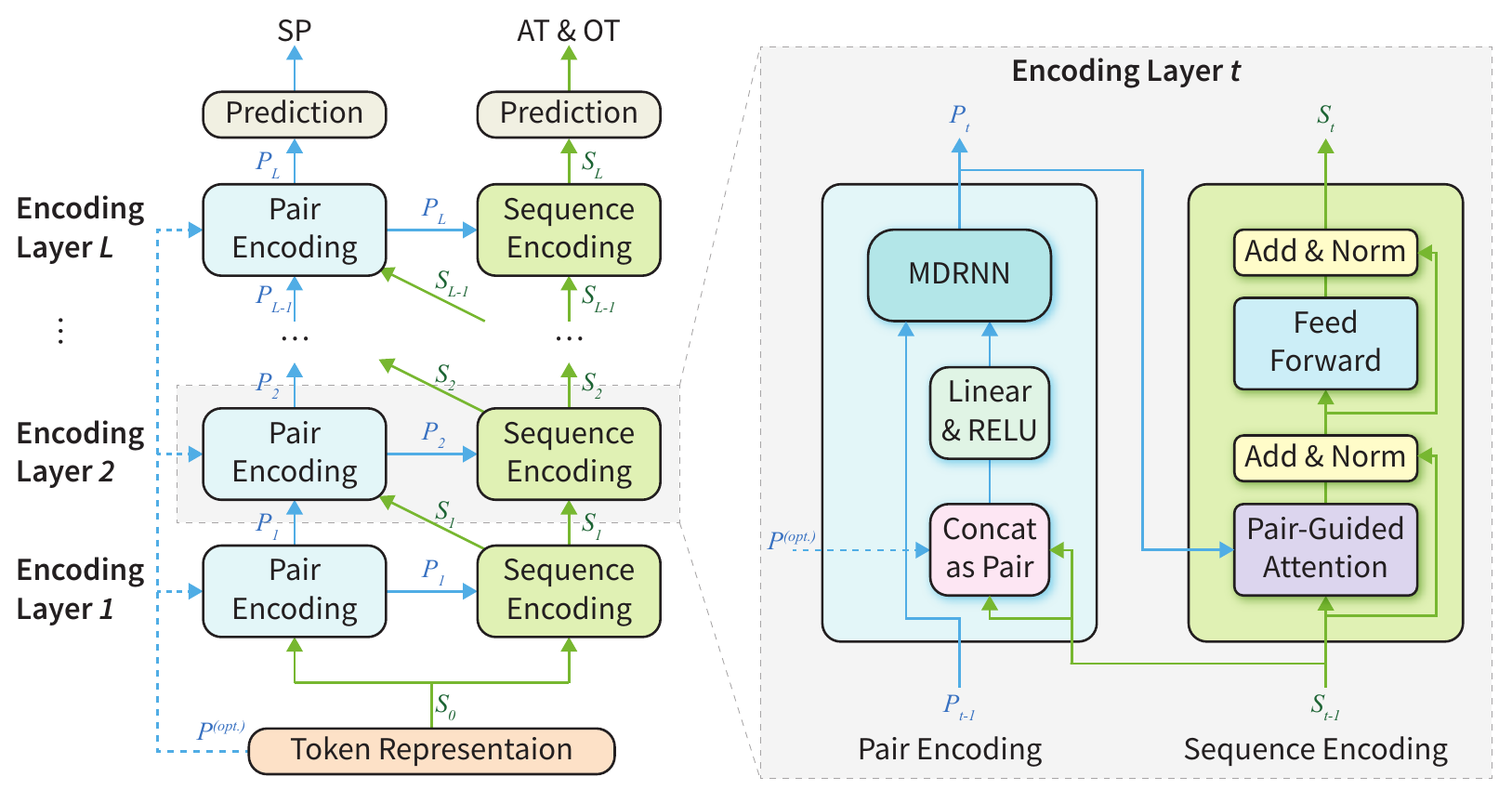}
    \caption{The framework of our model. Dashed lines are for optional components.}
    \label{fig:framework}
\end{figure*}

\section{Our Approach}

\subsection{Problem Formulation}

In this paper, we split the ABSA task into two periods: aspect/opinion term extraction and sentiment classification (SC), as shown in Figure \ref{fig:task_def}. 
The aspect/opinion term extraction subtask extracts the aspect terms (AT) and opinion terms (OT) in the sentences without considering the sentiment polarities (SP). Furthermore, according to the sentiment polarity tagging style of the dataset, the SC subtask is divided into two categories: ASTE, tagging SP on AT and OT, and AESC, which tags SP only on AT. 

In particular, we denote $\mathbf{AT}$, $\mathbf{OT}$ and $\mathbf{SP}$ as the set of predefined aspect terms, opinion terms and sentiment polarities, respectively, where 
$AT \in \mathbf{AT}$, $OT \in \mathbf{OT}$, and $SP \in \mathbf{SP}=$ \{POS, NEU, NEG\}. 
Given a sentence $s$ consisting of $n$ tokens ${\omega_1,\omega_2,...,\omega_n}$, we denote $T$ as the sentence output of our model. 
Specifically, for the ASTE task, $T=\{(AT,OT,SP)\}$, and for the AESC task, $T=\{(AT,SP)\}$.

\subsection{Model Overview}

Inspired by the work of \citet{wang-lu-2020-two} which utilize the dual-encoder structure, our approach for the ABSA task is designed to subtly modeling high affinity between aspect/opinion pair and ground truth by effectively leveraging the pair representation. 
As shown in Figure \ref{fig:framework}, our dual-encoder comprises two modules: (1) a sequence encoder, a Transformer network initialized with the pre-trained language model to represent AT and OT with the corresponding context. 
(2) a pair encoder, encoding the aspect-opinion pair (for ASTE) or aspect-aspect pair (for AESC) for each sentiment polarity.

\subsection{Token Representation}

For a length-$n$ input sentence $s= \omega_1, ... ,\omega_n$, besides the word-level representation $\mathbf{x}_{\text{word}}$, we also feed the characters of the word into the LSTM to generate the character-level representation $\mathbf{x}_{\text{char}}$.
Additionally, the pre-trained language model provides the contextualized representation $\mathbf{x}_{\text{plm}}$.
Finally, we concatenate these three representations of each word to feed into the dual-encoder:

\begin{equation}
    \mathbf{x}_i = [\mathbf{x}_{\text{char}};\mathbf{x}_{\text{word}};\mathbf{x}_{\text{plm}}].
\end{equation}

In our proposed dual-encoder architecture, we still treat the ASTE/AESC task as a unified sequence tagging task in previous work: for a given sentence $s$, where AT and OT on the main diagonal are annotated with B/I/O (Begin, Inside, Outside), each entry $E_{i,j}$ of the upper triangular matrix denotes the pair ($\omega_i$, $\omega_j$) from the input sentence. Our work is partially motivated by \citet{wu-etal-2020-grid} but significantly different. 

\begin{figure}[t]
    \centering
    \includegraphics[width=0.9\linewidth]{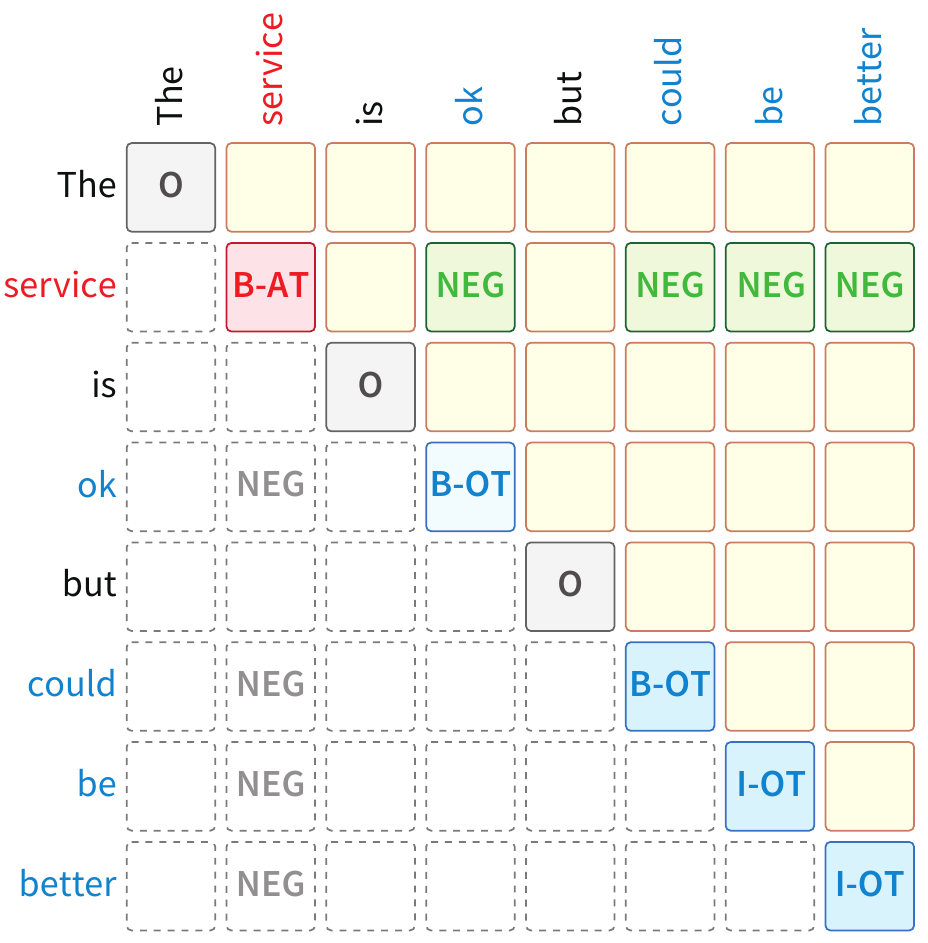}
    \caption{A tagging example for our model.}
    \label{fig:modeling}
\end{figure}

First, we improve the word-level pair representation to span-level pair representation with more accurate boundary information fed into our model. 
The tagging scheme of our model is illustrated in Figure \ref{fig:modeling}, in which the main diagonal are filled with AT and OT accompanying entries to the right of the main diagonal with span pairs. 
Compared to \cite{wu-etal-2020-grid}, our method may heavily reduce the redundancy aroused by AT and OT tags at the right of the main diagonal.

Second, we consider the context information on both two-dimension spaces and the historical information with the utilization of the recurrent neural network (RNN). 
However, \citet{wu-etal-2020-grid} merely adopted a single encoder which based on DE-CNN \cite{xu-etal-2018-double}/BiLSTM/BERT \cite{devlin-etal-2019-bert} to establish token representation, and they formulated the final word-pair representation by a simplified method of attention-guided word-pair concatenation. 

Thus, our dual-encoder could jointly encode AT, OT (with the corresponding context on both dimensions), and AT-OT pairs with representation information sharing.

\subsection{Sequence Encoder}

Following the previous work of \citet{DBLP:conf/nips/VaswaniSPUJGKP17}, we construct the sequence encoder as a Transformer network.

Here we apply a stack of $m$ self-attention layers, shown in Figure \ref{fig:framework}. Each layer consists of two sublayers: namely multi-head attention sublayer, feed-forward sublayer, at the top of each sublayer followed with both residual connection and layer normalization.

\subsubsection{Multi-head Attention Sublayer}

In this section, the token representation $\mathbf{x}_i$ is fed into a multi-head attention sublayer.

At first of our sequence encoder, the token representation $\mathbf{x}_i$ will be mapped into vector space as query $\mathbf{Q}_i$, key $\mathbf{K}_i$, value $\mathbf{V}_i$:
\begin{equation}
\begin{aligned}
    \mathbf{Q}_i&=\mathbf{x}_i \mathbf{W}_Q\\
    \mathbf{K}_i&=\mathbf{x}_i \mathbf{W}_K\\
    \mathbf{V}_i&=\mathbf{x}_i \mathbf{W}_V\\
\end{aligned}
\end{equation}
then the value vectors of all positions will be aggregated according to the normalized attention weight to get the single-head representation:

\begin{equation}
   \text{SingleHead} (\mathbf{Q}_i,\mathbf{K}_i,\mathbf{V}_i) =\text{softmax}( \frac{\mathbf{Q}_i\mathbf{K}_i^T}{\sqrt{d/m}})V
   \label{attention_main_formula}
\end{equation}
where $m$ is the number of heads, $d$ is the dimension of $\mathbf{x}_i$, and in our sequence encoder, $\mathbf{Q}=\mathbf{K}=\mathbf{V}=\mathbf{x}_i$. 

Then with multi-heads attention, our model builds up representations of the input sequence:
\begin{equation}
\begin{aligned}
    \mathbf{r}_i&=\text{MultiHead}(\mathbf{Q}_i,\mathbf{K}_i,\mathbf{V}_i)\\
    &=\text{Concat}(\text{SingleHead}_{1,..,m} (\mathbf{Q}_i,\mathbf{K}_i,\mathbf{V}_i)) \mathbf{W}^O\\  
\end{aligned}
\end{equation}
where $\mathbf{W}^O\in \mathbb{R}^d$.
We adopt the residual connection and layer normalization \cite{DBLP:journals/corr/BaKH16} on $\mathbf{r}_i$ and $\mathbf{x}_i$:
\begin{equation}
  \mathbf{a}_i= \text{LayerNorm}(\mathbf{r}_i+\mathbf{x}_i)
\end{equation}

\subsubsection{Feed-Forward Sublayer}
The outputs of the multi-head attention are fed into a feed-forward network:
\begin{equation}
     \mathbf{e}_i = \text{FFNN}(\mathbf{r}_i) = (\mathbf{a}_i\mathbf{W}_1+b_1)\mathbf{W}_2+b_2
\end{equation}
where $\mathbf{W}_1,\mathbf{W}_2,\in \mathbb{R}^{d\times d/m}$ and $b_1, b_2 \in \mathbb{R}^d$. 
At last, the sequence representation will be performed by layer normalization with residual connection:
 \begin{equation}
     \mathbf{S}_i=\text{LayerNorm}(\mathbf{e}_i+\mathbf{a}_i)
 \end{equation}

\subsection{Pair Encoder}

As shown in Eq. (\ref{attention_main_formula}), our task-specific pair representation is an $n\times n$ matrix of vectors, where the vector at row $i$ and column $j$ represents $i$-th and $j$-th word pair of the input sentence.
For the $l$-th layer of our network, we first add a Multi-Layer Perception (MLP) layer with ReLU \cite{DBLP:conf/icml/NairH10} to contextualize the concatenation of representations from the sequence encoder:
\begin{equation}
\mathbf{S}'_{l,i,j} = \text{ReLU}(\textbf{MLP} ([\mathbf{S}_{{l-1},i};\mathbf{S}_{l-1,j}]))
\end{equation}

Then we utilize the multi-dimensional recurrent neural network (MDRNN) \cite{graves2007multi} and gated recurrent unit (GRU) \cite{cho-etal-2014-learning} to contextualize $\mathbf{S}'_{l,i,j}$.
The contextualized pair representation $\mathbf{P}_i$ is computed iteratively from the hidden states of each cell:
\begin{equation}
    \mathbf{P}_{l,i,j}=\text{GRU}(
    \mathbf{S}'_{l,i,j}, 
    \mathbf{P}_{l-1,i,j}, 
    \mathbf{P}_{l,i-1,j}, 
    \mathbf{P}_{l,i,j-1})
\end{equation}
The pair encoder does not consider only the word pair at neighboring rows and columns but also those of the previous layer.

\subsection{Training}

Given a sentence $s$ with pre-defined tags $AT$, $OT$, and $SP\in$ \{POS, NEU, NEG\}, we denote the AT or OT tag of token $\omega_i$ as $a_i$ and the SP tag between the tokens $\omega_i$ and $\omega_j$ as $t_{ij}$. 
To predict the label of the posterior of the aspect/opinion terms $\hat{y}$, we apply a softmax layer on the sequence embedding of aspect/opinion terms $\mathbf{S}_l$. Similarly, to obtain the distribution of sentiment polarity type label $\hat{v}$, we apply softmax on the pair representation of $\mathbf{P}_l$:
\begin{equation}
    P(\hat{y}|a_i,s)={\text{softmax}} (\mathbf{W}_{term}\mathbf{S}_l)
\end{equation}
\begin{equation}
    P(\hat{v}|t_{ij},s)={\text{softmax}} (\mathbf{W}_{pola}\mathbf{P}_l)
\end{equation}
where $\mathbf{W}_{term}$ and $\mathbf{W}_{pola}$ are learnable parameters.  

At the training, we adopt the Cross-Entropy as our loss function. For the gold aspect and opinion term $a_i\in \mathbf{AT} \bigcap \mathbf{OT}$ and gold polarity $t_{ij}\in \mathbf{SP}$, the training losses are respectively:
\begin{equation}
    \mathcal{L}_{term} =-\sum_{a_i \in \mathbf{AT} \cap \mathbf{OT}} \log(P(\hat{y}=y|a_i,s))
\end{equation}

\begin{equation}
    \mathcal{L}_{pola} =-\sum_{t_{ij} \in \mathbf{SP},i \neq j} \log(P(\hat{v}=v|t_{ij},s))
\end{equation}
where the $y$ and $v$ are the gold annotations of corresponding terms.

To jointly train the model, we utilize the summation of these two loss functions as our training objective:
\begin{equation}
   \mathcal{L}=\mathcal{L}_{term}+\mathcal{L}_{pola}
\end{equation}

\begin{table*}[t]
    \centering
    \small
    \label{main_results}
    \resizebox{1.0\textwidth}{!}{
    \begin{tabular}{lccc|ccc|ccc|ccc}
        \toprule
        \multirow{2}{*}{ \textbf{Models}} & \multicolumn{3}{c}{\it14Rest} & \multicolumn{3}{c}{\it14Lap} & \multicolumn{3}{c}{\it15Rest}& \multicolumn{3}{c}{\it16Rest} \\
        \cmidrule(r){2-4} \cmidrule(r){5-7} \cmidrule(r){8-10}\cmidrule(r){11-13}
         &  \it P. &  \it R.   &   \it $F_1$ &  \it P. &  \it R.   &   \it $F_1$ &  \it P. &  \it R.   &   \it $F_1$&  \it P. &  \it R.   &   \it $F_1$\\
        \midrule
        CMLA$\text{+}$ &39.18 & 47.13   & 42.79  &30.09&36.92&33.16&34.56&39.84&37.01& 41.34  & 42.10& 41.72    \\
        RINANTE$\text{+}$&31.42& 39.38 & 34.95  &21.71&18.66&20.07 &29.88&30.06&29.97   & 25.68 & 22.30& 23.87   \\
        Li-unified-R&41.04  & 67.35  & 51.00&40.56&44.28&42.34&44.72&51.39&47.82    & 37.33  & 54.51 & 44.31      \\
        \cite{DBLP:conf/aaai/PengXBHLS20}&43.24 &  63.66 &  51.46 &37.38&50.38&42.87&48.07&57.51&52.32            & 46.96   & 64.24  &  54.21       \\
        OTE-MTL&63.07& 58.25 &60.56&54.26&41.07&46.75&60.88&42.68&50.18 & 65.65& 54.28& 59.42            \\
        GTS-BiLSTM&71.41  &53.00&60.84  & 58.02 &40.11 &47.43  &64.57 &44.33 &52.57                  & 70.17  & 55.95 & 62.26 \\
        ${\text{JET}^\text{t}}$&66.76  &49.09   &56.58    &52.00&35.91&42.48&59.77&42.27&49.52 & 63.59 & 50.97 &  56.59       \\
        ${\text{JET}^\text{o}}$&61.50  & 55.13 &58.14&53.03&33.89&41.35&64.37&44.33&52.50 & 70.94  & 57.00 &  63.21\\
        \midrule
       ${\text{GTS}_{\text{+}\text{BERT}}}$   & 71.76    & 59.09          & 64.81 &57.12&53.42&55.21&54.71&55.05&54.88&65.89 & 66.27 & 66.08     \\
       ${\text{JET}^\text{t}_{\text{+}\text{BERT}}}$  &63.44  &54.12   &58.41      &53.53&43.28&47.86&\bf68.20&42.89&52.66  & 65.28  & 51.95& 57.85  \\
        {${\text{JET}^\text{o}_{\text{+}\text{BERT}}}$} &70.56  &55.94   &62.40      &55.39&47.33&51.04&64.45&51.96&57.53  & 70.42  & 58.37& 63.83  \\
         
        {${\text{\cite{huang2021first}}_{\text{+}\text{BERT}}}$} &63.59 & 73.44   &  68.16 &57.84&59.33&  58.58 & 54.53 & 63.30 & 58.59 & 63.57  & 71.98&  67.52 \\
        
         \midrule
          {${\text{Ours}_{\text{+}\text{BERT}}}$}  &67.95 & 71.23   &  69.55 &62.12&56.38& 59.11 & 58.55 & 60.00 & 59.27 & 70.65  & 70.23& 70.44 \\
         
       \bf {${\text{Ours}_{\text{+}\text{ALBERT}}}$}  &\bf75.20 & \bf74.45   &  \bf74.82 &\bf66.67&\bf60.26& \bf 63.30 & 66.74 & \bf69.69 & \bf 67.67 & \bf71.40  & \bf74.32& \bf 72.01 \\
        \bottomrule
    \end{tabular}}
    \caption{Results on \emph{ASTE-Data-V2} test datasets. Baseline results are directly retrieved from \cite{huang2021first}. The extensive experiment of \emph{ASTE-Data-V1} test datasets are supplemented in the Appendix.}
    \label{mainresults}
\end{table*} 

\section{Experiments}
\subsection{Data}

To make a fair comparison with previous methods, we adopt two versions of datasets for the ASTE task: (1) \emph{ASTE-Data-V1}, originally provided by \citet{DBLP:conf/aaai/PengXBHLS20} from the SemEval 2014 Task 4 \cite{pontiki-etal-2014-semeval}, SemEval 2015 Task 12 \cite{pontiki-etal-2015-semeval} and SemEval 2016 Task 5 \cite{pontiki-etal-2016-semeval}, and (2) \emph{ASTE-Data-V2}, the refined version annotated by \citet{xu-etal-2020-position}, with additional annotation of implicitly overlapping triplets. Furthermore, the name of each dataset is composed of two parts. The former part denotes the year when the corresponding SemEval data was proposed, and the latter part is the domain name of the reviews on restaurant service or laptop sales. Data statistics of them is shown in Table \ref{dataset_statistic_main}.

Then, for the AESC task, we adopt the dataset annotated by \citet{DBLP:conf/aaai/WangPDX17}, which is composed of three datasets, and the statistics is shown in Table \ref{dataset_statistic_racl}.
The implementation details of our dual-encoder model are unfolded in Appendix \ref{Imple_details} for the sake of putting main concentration on our argument. Our code will be available at \url{https://github.com/Betahj/PairABSA}.

\subsection{Results on the ASTE Task}\label{section:3.2}

Our model will compare to the following baselines on the ASTE task, and more details about these baseline models are listed in Appendix \ref{baseline_intro}.

1) \textbf{RINANTE+} \cite{DBLP:conf/aaai/PengXBHLS20}. 

2) \textbf{CMLA+} \cite{DBLP:conf/aaai/PengXBHLS20}. 

3) \textbf{Li-unified-R} \cite{DBLP:conf/aaai/PengXBHLS20}.

4) \textbf{Peng et al.} \cite{DBLP:conf/aaai/PengXBHLS20}. 

5) \textbf{OTE-MTL} \cite{zhang-etal-2020-multi-task}. 

6) \textbf{JET} \cite{xu-etal-2020-position}.

7) \textbf{GTS} \cite{wu-etal-2020-grid}. 

8) \textbf{Huang et al.} \cite{huang2021first}.

 \begin{table*}[t]
    \centering
    \small
    \label{AESC_results}
    \begin{tabular}{cccc|ccc|ccc}
    \toprule
    \multirow{2}{*}{\bf Models} & \multicolumn{3}{c}{\it14Rest} & \multicolumn{3}{c}{\it14Lap} & \multicolumn{3}{c}{\it15Rest} \\
    \cmidrule(r){2-4} \cmidrule(r){5-7} \cmidrule(r){8-10}
     &  \it AE &  \it OE   &   \it AESC &  \it AE &  \it OE   &   \it AESC &  \it AE &  \it OE   &   \it AESC  \\
    \midrule
    SPAN-BERT &86.71 & -   & 73.68  &82.34&-&61.25&74.63&-&62.29  \\
    IMN-BERT&84.06& 85.10 & 70.72  &77.55&81.00&61.73 &69.90&73.29&60.22   \\
    RACL-BERT & 86.38  & 87.18&75.42&81.79&79.72&63.40&73.99&76.00    & 66.05      \\
    \cite{mao2021joint}&86.60 &  - &75.95&82.51&-&65.94&75.08&-  & 65.08         \\
    \midrule
     Baseline$_{\text{+}\text{BERT}}$  & 86.64  & 85.59 & 70.20  &  80.03 & 80.52 & 57.81 & 72.24 & 75.72 & 62.91 \\
     {${\text{Ours}_{\text{+}\text{BERT}}}$}  
     &86.94 & 85.80  &  70.49 & 80.26 & 80.61 & 57.98  & 72.68 & 75.94&63.19 \\
    \bf{${\text{Ours}_{\text{+}\text{ALBERT}}}$} 
    &  86.52 &  85.82  &  74.19 &  81.80 &  80.47 &   61.51 &  \bf75.42 & \bf 78.86&64.82 \\
    \bottomrule
    \end{tabular}
    \caption{Results for AESC on the test datasets annotated by \citet{DBLP:conf/aaai/WangPDX17}. Baseline results are directly retrieved from \cite{mao2021joint}. The best result of each evaluation metric is bolded.}
    \label{AESC_mainresults}
\end{table*} 

The main results of all the models on the ASTE task are shown in Table \ref{mainresults}. 
Compared with the best baseline model \cite{huang2021first}, our BERT-based dual-encoder model achieves an improvement by 1.39, 0.53, 0.68, and 2.92 absolute \text{$F_1$} score on benchmark datasets. 
This result signifies that our dual-encoder model is capable of capturing the difference between AT/OT extraction subtask and SC subtask with the help of the additional pair encoder.    
Besides, our ALBERT-based model significantly outperforms all the other competitive methods on most metrics of 4 datasets {\it14Rest}, {\it14Lap}, {\it15Rest} and {\it16Rest} except for precision score of {\it15Rest}. Most notably, our ALBERT-based model achieves an improvement of 6.66, 4.72, 9.08, and 4.49 absolute \text{$F_1$} score over all the baseline models on four benchmark datasets, respectively. This result demonstrates the superiority of our dual-encoder model. However, we notice that our precision score of {\it15Rest} is comparable to \cite{xu-etal-2020-position}, which might be due to our model is more biased towards positive predictions but that the F1 score still suggests it is an overall improvement. 

The similar phenomenon that our BERT-based dual-encoder model shows larger improvements in F1 scores on 14Rest (1.39) and 16Rest (2.92) than on 14Lap (0.53) and 15Rest (0.68) verifies the explanation of \citet{xu-etal-2020-position} on large distribution differences of {\it14Rest} and {\it15Rest}.
Nevertheless, we also observe a different phenomenon that our ALBERT-based dual-encoder model achieves significant $F_1$ score improvements on {\it14Rest} (6.66) and {\it15Rest} (9.08), better than {\it14Lap} (4.72) and {\it16Rest} (4.49), makes a challenge to the explanation developed by \citet{xu-etal-2020-position}.
From our perspective, it might be due to the different fitting degree between the distribution of \emph{ASTE-Data-V2} datasets and corresponding pre-trained language models.
Additionally, we evaluate our model on the \emph{ASTE-Data-V1} datasets and then experimental results further demonstrate the effectiveness of our dual-encoder model. These results are shown in Table \ref{aste_v1_results} of the Appendix.

\subsection{Results on the AESC Task}

For the AESC task, our model will compare to the following baselines:
 
1) \textbf{SPAN-BERT} \cite{hu-etal-2019-open}. 
 
2) \textbf{IMN-BERT} \cite{hu-etal-2019-open}. 

3) \textbf{RACL-BERT} \cite{chen-qian-2020-relation}. 

4) \textbf{Mao et al.} \cite{mao2021joint}.

To investigate whether the performance of our model on the AESC task maintains the same efficiency as the ASTE task, we conduct a series of experiments on AESC datasets. Results of all the models on the AESC task are shown in Table \ref{AESC_mainresults}. 
Compared with the best baseline model of \citet{mao2021joint}, the performance of our model is not comparable except for the absolute \text{$F_1$} score on AE and OE of {\it 15Rest} dataset. 
Then, to excavate the contribution of our dual-encoder structure on the AESC task, we evaluate our model on the baseline without the pair encoder. From Table \ref{AESC_mainresults} we can see that the performance of our dual-encoder model is comparable on the AESC task than single-encoder structure. The AESC task is only a simplified version of the ASTE task without taking 
AE/OE paring and sentiment polarity classification into consideration reversely, which is the training objective of our joint model with the help of task-specific structure design. Consequently, our model is incapable of functioning well in the AESC task.

\section{Ablation Studies}

\subsection{Different Pre-trained Language Models}
We conduct the experiment on the {\it14Lap} of \emph{ASTE-Data-V2} datasets to excavate the performance of three frequently utilized pre-trained language models (PLMs): XLNet \cite{DBLP:conf/nips/YangDYCSL19}, RoBERTa \cite{DBLP:journals/corr/abs-1907-11692} and ALBERT \cite{DBLP:conf/iclr/LanCGGSS20}. 

Table \ref{plmperformance_comparison} shows that ALBERT helps achieve the best result among these four PLMs. 
However, even with BERT as the baseline model \cite{xu-etal-2020-position,huang2021first}, our model also performs better. 
We also notice that, different from most models, our model is sensitive to different PLMs. Specifically, the absolute \text{$F_1$} score between BERT and RoBERTa, ALBERT is 3.90 and 7.05, respectively. It demonstrates that our model performance could effectively be boosted by our choice of PLM, and thus we choose ALBERT as our base encoder. 

\begin{table}[ht]
    \centering
    \small
    \begin{tabular}{cccc}
    \toprule
    \multirow{1}{*}{ PLM} &  \it P.&\it R.&{$ F_1$}\\
	\midrule
	
	XLNet &63.24 &51.20 &56.59 \\
	BERT	&62.12 &56.38 &59.11\\
	RoBERTa	&61.79 &58.60 &60.15\\
    ALBERT	&66.67 &60.26 &63.30 \\
	\bottomrule
\end{tabular}
\caption{Comparison of our model with different pre-trained language models on {\it14Lap} test set of \emph{ASTE-Data-V2}.}
\label{plmperformance_comparison}
\end{table}

\subsection{Dual-encoder Structure}
Therefore, the joint modeling method must take not only the fitting degree between individual modules and subtasks but also the difference of each module into consideration.

\begin{table}[ht]
    \centering
    \small
    
    \begin{tabular}{cccc}
    \toprule
    \multirow{1}{*}{Settings} &  {\it P.}&{\it R.}&{$ F_1$}\\
	\midrule
		Default Setting	&66.67 &60.26 &63.30\\
	w/o Pair Encoder&58.16 &59.15 &58.65 \\
	w/o Interaction &64.55 & 58.88 & 61.58 \\
	\bottomrule
\end{tabular}
\caption{Ablation of our dual-encoder structure on {\it14Lap} test set of \emph{ASTE-Data-V2}.}
\label{dual-encoder_structure_comparison}
\end{table}

\subsection{Number of Encoder Layers} 

\begin{figure}[t]
	\centering
		\setlength{\abovecaptionskip}{0pt}
		\begin{center}	\pgfplotsset{height=5.9cm,width=8.5cm,compat=1.14,every axis/.append style={thick},every axis legend/.append style={ at={(1,1)}},legend columns=1}
			\begin{tikzpicture}
			\tikzset{every node}=[font=\small]
			\begin{axis}
			[width=7cm,enlargelimits=0.15, tick align=outside, xticklabels={ $1$, $2$, $3$, $4$, $5$, $6$, $7$, $8$, $9$, $10$},
            axis y line*=left,
            xtick={0,1,2,3,4,5,6,7,8,9},
            ylabel={$F_1$ score},
            axis x line*=left, 
            ymax=75,
            ylabel style={align=left},xlabel={Number of Layers},font=\small]
			\addplot+ [sharp plot,mark=square*,mark size=1.2pt,mark options={solid,mark color=orange}, color=blue] coordinates
			{ (0,60.66)(1,60.67)(2,63.30)(3,61.88)(4,61.38)(5,61.21)(6,59.78)};
			\addlegendentry{\tiny{\it14Lap}}
            \addplot+ [sharp plot,densely dashed,mark=triangle*,mark size=1.2pt,mark options={solid,mark color=orange}, color=orange] coordinates
			{ (0,69.4)(1,70.28)(2,72.01)(3,70.91)(4,71.80)(5,68.85)(6,71.12)};
			\addlegendentry{\tiny \it16Rest}
            \end{axis}
			\end{tikzpicture}
		\end{center}
	\caption{The impact of number of encoder layers on model performance.}\label{numlayer} 
\end{figure}
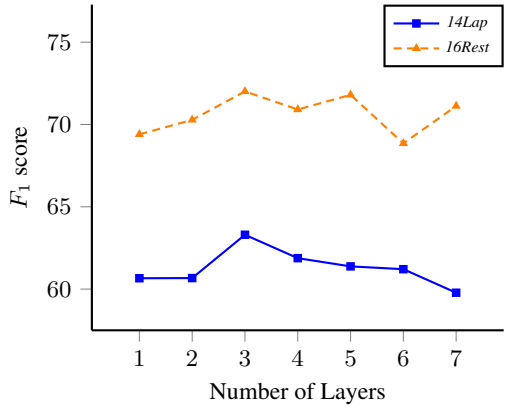

The results with different numbers of encoder layers are in Figure \ref{numlayer}. Generally, the performance of triplet extraction synchronously increases with the number of encoder layers of both dataset distributions. Nevertheless, when the number of encoder layers exceeds 3, the performance shows a continuous decreasing trend, except that on {\it16Rest} when the number of encoder layers is increased to 7, the performance increases by nearly 2.5 absolute \text{$F_1$} score. Despite this inconsistent phenomenon, to mainly consider computational/time complexities, we adopt 3 as the number of encoders. 

\subsection{The Impact of The Number of GRU}

Table \ref{GRU_dimension_performance_comparison} shows the results with different settings of multi-dimensional recurrent neural networks. The \textit{Uni-directional} denotes the hidden state from forward GRU results in one quadrant of same dimension space, the \textit{Bi-directional} denotes the hidden state from forward and backward GRU results in two quadrants of same dimension space, and \textit{Quad-directional} denotes the hidden state from forward and backward GRU results in four quadrants of same dimension space. We observe that the \textit{Quad-directional} setting significantly outperforms the other two settings. 
It is also noteworthy that the performance gap between \textit{Bi-directional} and \textit{Uni-directional} dimensions is much lower than the gap between \textit{Quad-directional} and \textit{Bi-directional} dimensions, which might be the reason why most previous work using bidirectional modelings cannot perform well. 
Thus, we choose \textit{Quad-directional} as the dimensional setting of our multi-dimensional RNNs. 

\begin{table}[t]
    \centering
    \small
    \begin{tabular}{cccc}
    \toprule
    \multirow{1}{*}{Settings} &  {\it P.}&{\it R.}&{$ F_1$}\\
	\midrule
	Uni-directional	    &63.51 &59.52 &61.45\\
	Bi-directional      &64.96 &58.60 &61.61 \\
	Quad-directional	&66.67 &60.26 &63.30\\
	\bottomrule
    \end{tabular}
    \caption{Ablation of different settings of multi-dimensional recurrent neural networks on {\it14Lap} test set of \emph{ASTE-Data-V2}.}\label{GRU_dimension_performance_comparison}
\end{table}

\subsection{The Effect of Character-level Representation}

\begin{table*}[t] 
    \begin{center}  
    \scalebox{0.8}{
    \begin{tabular}{lp{17cm} }  
    \toprule
     
     \textbf{Example-1}&Also stunning colors and speedy.\\
     \midrule
    \textbf{gold} 
    &Also [stunning]$\scriptstyle{\rm_{{OT}|POS_{t_1}}}$ [colors]$\scriptstyle{\rm_{{AT}|POS_{t_1}}}$ and speedy.\\
    \midrule
    \textbf{predict}
     & Also \textcolor{blue}{[stunning]}$\scriptstyle{\rm_{{\color{blue}{OT}}|{{\color{black}POS_{t_1}}}|{\color{black}POS_{t_2}}}}$ \textcolor{red}{[colors]}$\scriptstyle{\rm_{{{\color{red}{AT}}}|POS_{t_1}}}$ and \textcolor{red}{[speedy]}$\scriptstyle{\rm_{{{\color{red}{AT}}}|POS_{t_2}}}$.\\
     \midrule
     
    \textbf{Example-2}&Excellent performance, usability, presentation and time response.\\
     \midrule
    \textbf{gold} & [Excellent]$\scriptstyle{\rm_{{OT}|POS_{t_1}|POS_{t_2}|POS_{t_3}|{POS_{t_4}}}}$ [performance]$\scriptstyle{\rm_{{AT}|POS_{h_1}}}$, [usability]$\scriptstyle{\rm_{{AT}|POS_{h_2}}}$, [presentation]$\scriptstyle{\rm_{{AT}|POS_{h_3}}}$ and [time response]$\scriptstyle{\rm_{{AT}|POS_{h_4}}}$.\\
    \midrule
    \textbf{predict} &\textcolor{blue}{[Excellent]}$\scriptstyle{\rm_{{\color{blue}{OT}}|{{\color{black}POS_{t_1}}}|{\color{black}POS_{t_2}}|{\color{black}POS_{t_3}}|{\color{black}POS_{t_4}}}}$ 
    \textcolor{red}{[performance]}$\scriptstyle{\rm_{{\color{red}{AT}}|{\color{black}POS_{h_1}}}}$,
    \textcolor{red}{[usability]}$\scriptstyle{\rm_{{\color{red}{AT}}|{\color{black}POS_{h_2}}}}$, 
    \textcolor{red}{[presentation]}$\scriptstyle{\rm_{{\color{red}{AT}}|{\color{black}POS_{h_3}}}}$ and 
    \textcolor{red}{[time response]}$\scriptstyle{\rm_{{\color{red}{AT}}|{\color{black}POS_{h_4}}}}$.\\
      
    \midrule
    \textbf{Example-3}&OSX Lion is a great performer..extremely fast and reliable.\\
    \midrule
    \textbf{gold}&[OSX Lion]$\scriptstyle{\rm_{{AT}|POS_{h_1}|POS_{h_2}|POS_{h_3}}}$is a [great]$\scriptstyle{\rm_{{OT}|POS_{t_1}}}$ performer..extremely [fast]$\scriptstyle{\rm_{{OT}|POS_{t_2}}}$ and  [reliable]$\scriptstyle{\rm_{{OT}|POS_{t_3}}}$.\\
    \midrule
       \textbf{predict}&\textcolor{red}{[OSX Lion]}$\scriptstyle{\rm_{{\color{red}{AT}}|{POS_{h_1}}|{POS_{h_2}}|{POS_{h_3}}}}$is a \textcolor{blue}{[great]}$\scriptstyle{\rm_{{\color{blue}{OT}}|POS_{t_1}}}$ performer..extremely \textcolor{blue}{[fast]}$\scriptstyle{\rm_{{\color{blue}{OT}}|{POS_{t_2}}}}$ and \textcolor{blue}{[reliable]}$\scriptstyle{\rm_{{\color{blue}{OT}}|POS_{t_3}}}$.\\
     \midrule
      
       \textbf{Example-4}&I am please with the products ease of use; out of the box ready; appearance and functionality.\\
      \midrule
      \textbf{gold}&I am [please]$\scriptstyle{\rm_{{OT}|POS_{t_1}|POS_{t_2}|POS_{t_3}}}$ with the products [ease]$\scriptstyle{\rm_{{OT}|POS_{t_4}}}$ of [use]$\scriptstyle{\rm_{{AT}|POS_{h_1}|POS_{h_4}}}$; out of the box ready;\newline [appearance]$\scriptstyle{\rm_{{AT}|POS_{h_2}}}$ and [functionality]$\scriptstyle{\rm_{{AT}|POS_{h_3}}}$.\\
      
      \midrule
      \textbf{predict}&I am \textcolor{blue}{[please]}$\scriptstyle{\rm_{{\color{blue}{OT}}|{\color{teal}{POS_{t_1}}}|{\color{orange}POS_{t_2}}|{\color{violet}POS_{t_3}}}}$ with the products \textcolor{blue}{[ease]}$\scriptstyle{\rm_{{\color{blue}{OT}}|{\color{olive}POS_{t_4}}}}$ of \textcolor{red}{[use]}$\scriptstyle{\rm_{{\color{red}{AT}}|{\color{teal}POS_{h_1}}|{\color{olive}POS_{h_4}}}}$; out of the box ready;\newline \textcolor{red}{[appearance]}$\scriptstyle{\rm_{{\color{red}{AT}}|{\color{orange}POS_{h_2}}}}$ and \textcolor{red}{[functionality]}$\scriptstyle{\rm_{{\color{red}{AT}}|{\color{
      violet}POS_{h_3}}}}$.\\
    \bottomrule 
    \end{tabular}}
    \end{center} 
  \caption{Case study of our proposed model, where $\rm AT/OT$  denote aspect term/opinion term, $\rm POS$ denotes sensitive polarity of positive, the subscript of sensitive polarity $\rm h_1/t_1$ denotes the head/tail term of the 1st pair in terms of corresponding sentiment, etc.} 
  \label{casestudy}
\end{table*}

To investigate the contribution of character-level representation to our input sequence, we remove the character-level representation generated by LSTM. Experimental result shows that the performance decreases by 0.44 absolute \text{$F_1$} score.

\section{Case Study}

To investigate why our model far exceeds the baseline models, we conduct a case study of three typical cases from {\it14Lap} test dataset of \emph{ASTE-Data-V1}, as shown in Table \ref{casestudy}.

From \textbf{Example-1}, we observe that our model is able to handle the one-to-one case. However, our dual-encoder structure is more biased towards coordinative relation between \emph{colors} and \emph{speedy}. More cases we investigated further demonstrating that our model performs slightly worse on on-to-one than one-to-many and many-to-many relation types.
From \textbf{Example-2}, we see that our model can tackle the one-opinion to many-target problem. However, most previous works are even unable to tackle one-opinion to two-target. 
From \textbf{Example-3}, we observe that our model is capable of well handling the one-target to many-opinion problem, which is neglected by most of the existing work but important for triplet extraction. 
Because many sentences compose conflicting sentiments on target, the model will fail to recognize the opposite polarity of the same AT when the incorrect AT extraction happens. 
Finally, we also observe that our model accurately inferences the boundary of \emph{OSX Lion} span, which demonstrates the usefulness of our transformation that utilizes span to replace the word.
From \textbf{Example-4}, we notice that our model could efficiently handle the complex situation of many-opinion to many-target with long-range dependency, which was particularly paid attention to but not solved well by \citet{zhang-etal-2020-multi-task}. 
It is due to incorporating the self-attention mechanism and GRU in two dimensions, and our model is sensitive to the difference between the proposed dual-encoder architecture.
Collectively, these aforementioned cases demonstrate the robustness of our dual-encoder model.

\section{Related Work}

Recently, NLP has been developed rapidly~\cite{he-etal-2018-syntax,li-etal-2018-seq2seq,cai-etal-2018-full,DBLP:conf/aaai/LiHZZZZZ19,DBLP:journals/corr/abs-2009-08775,DBLP:conf/aaai/0001YZ21}, and the process is further by deep neural networks~\cite{parnow-etal-2021-grammatical,li2021text} and pre-trained language models~\cite{10.1162/coli_a_00408,DBLP:conf/aaai/0001WZLZZZ20}. Aspect-based sentiment analysis was proposed by \citet{pontiki-etal-2014-semeval} and also received lots of attention in recent years.

\subsection{ASTE Task}

The ASTE task aims to make triplet extraction of aspect terms, opinion terms, and sentiment polarity, which was introduced by \citet{DBLP:conf/aaai/PengXBHLS20}. 
In their work, they leveraged the sequence labeling method to extract aspect terms and target sentiment and utilized graph neural networks to detect candidate opinion terms.  
\citet{zhang-etal-2020-multi-task} proposed a multi-task framework that decomposes the original ASTE task into two subtasks, sequence tagging of AT/OT, and word pair dependency parsing.
For joint learning, \citet{xu-etal-2020-position} proposed a sequence tagging framework based on LSTM-CRF. \citet{wu-etal-2020-grid} constructed an encoder-decoder model to handle this task with grid representation of aspect-opinion pairs.
Then with the incorporation of a more specific semantic information guide for the proposed model, the ASTE is transformed as MRC task \cite{chen2021bidirectional,mao2021joint}.
Recently, \citet{huang2021first} proposed a sequence tagging-based model to perform representation learning on the ASTE task.

\subsection{AESC Task}

The AESC task is to perform aspect terms extraction and sentiment classification simultaneously. 
\citet{hu-etal-2019-open} and \citet{DBLP:conf/ijcai/ZhouHGHH19} used a span-level sequence tagging method to tackle huge search space and sentiment inconsistency problems. 
Although the huge search space issue has been solved by \citet{hu-etal-2019-open}, there still exists a low-performance problem. 
Addressing this issue, \citet{lin-yang-2020-shared} utilized a BERT encoder to contextualize shared information of target extraction and target classification subtasks. Meanwhile, they used two BiLSTM networks to encode the private information of each subtask, which greatly boosted the model performance.

\subsection{Dual-encoder Structure}

Productive efforts were put into the research of dual-encoder structure for natural language processing tasks in the last few years because of the natural ability to model representational similarity maximization associated tasks \cite{chidambaram-etal-2019-learning,DBLP:conf/aaai/YuLGZW20,DBLP:journals/corr/abs-2103-05028}. Generally, these approaches encoded a single component of the approaches encoded a single component of the corresponding task separately for the processing in the next phase. Recently, \citet{wang-lu-2020-two} proposed a sequence-table representation learning architecture for a typical triplet extraction task: relation extraction, and this work established an example of tacking the triplet extraction task with the dual-encoder based architecture.

\section{Conclusion}

In this paper, we observe the significant differences between the AT/OT extraction subtask and the SC subtask of ABSA for the joint model.
Specifically, the results on 8 benchmark datasets with significant improvement over state-of-the-art baselines verify the effectiveness of our proposed model. Furthermore, to distinguish such differences and keep the shared part between different modules simultaneously, we construct a dual-encoder framework with representation learning and self-attention mechanism. In addition to the encoder-sharing approach, our dual-encoder framework can capture the difference between the subtasks by interconnecting encoders at each layer to share the critical information. 

\section{Acknowledgement}
We appreciate \citeauthor{wang-lu-2020-two} for their provided open resource. Based on this, we conducted our work on ABSA. We also appreciate the help from the reviewers and program chairs.

\bibliography{emnlp2021}
\bibliographystyle{acl_natbib}

\clearpage

\appendix

\section{Additional Results}

\subsection{Evaluation Metric}

We adopt $F_1$ score as our evaluation metric as other baseline models. In precise, we measure the F1 score calculated between the final exact match of AT/OT span, AT/OT types and corresponding polarity predictions and gold triplets.

\subsection{Implementation Details} \label{Imple_details}

For the token representation, we utilize 100-dimensional GloVe \cite{pennington-etal-2014-glove} as initialization and restrict the update of word embedding. 
The hidden size is 200. 
The decay rate is 0.05, and the decay steps are 1000. 
Besides, to further boost the performance of our proposed model, we utilize the {\tt ALBERT-xxlarge-v1} \cite{DBLP:conf/iclr/LanCGGSS20} as our pre-trained language model. 
We also use Adam with a learning rate of 0.001 and update parameters with a batch size of 24. 
Training is limited to the preset max steps. 
All models are implemented on the TITAN RTX. 
More implementation details are listed in Table \ref{parameter_settings}.

\begin{table}[h]
    \centering
    \small
    \begin{tabular}{lr} 
    \toprule  
    \textbf{Setting} & \textbf{Value} \\  
    \midrule
    Char/    Char/Word/Glove  &     100 \\  
    Word/Glove  &     100 \\  
    Hidden Embedding Dim &200\\
    Token Embedding Dim &100\\
    Char Embedding Dim &30\\
    Gradient Clipping &5.0\\
    Batch Size&24\\
    Optimizer&Adam\\
    Learning Rate  & $1e^{-3}$ \\
    Dropout Rate & 0.5 \\
    Decay Rate & 0.05 \\
    Number of Layer& 3\\
    Attention Heads &8\\
    \bottomrule
    \end{tabular} 
    \caption{Hyperparameter settings for our models}
    \label{parameter_settings}
\end{table}

\begin{table*}[t]
\centering
\small
\label{main_v1_results}
    \resizebox{1.0\textwidth}{!}{
    \begin{tabular}{lccc|ccc|ccc|ccc}
        \toprule
        \multirow{2}{*}{\bf Models} & \multicolumn{3}{c}{\it14Rest} & \multicolumn{3}{c}{\it14Lap} & \multicolumn{3}{c}{\it15Rest}& \multicolumn{3}{c}{\it16Rest} \\
        \cmidrule(r){2-4} \cmidrule(r){5-7} \cmidrule(r){8-10}\cmidrule(r){11-13}
         &  \it P. &  \it R.   &   \it {$\rm F_1$} &  \it P. &  \it R.   &   \it {$\rm F_1$} &  \it P. &  \it R.   &   \it {$\rm F_1$}&  \it P. &  \it R.   &   \it {$\rm F_1$}\\
        \midrule
        CMLA$\text{+}$ &40.11 & 46.63   & 43.12  &31.40&34.60&32.90&34.40&37.60&35.90& 43.60  & 39.80& 41.60    \\
        RINANTE$\text{+}$&31.07& 37.63 & 34.03  &23.10&17.70&20.00 &29.40&26.90&28.00   & 27.10 & 20.50& 23.30   \\
        Li-unified-R&41.44  & 68.79  & 51.68&42.25&42.78&42.47&43.34&50.73&46.69    & 38.19  & 53.47 & 44.51      \\
        \cite{DBLP:conf/aaai/PengXBHLS20}&44.18 &  62.99 &  51.89 &40.40&47.24&43.50&40.97&54.68&46.79            & 46.76   & 62.97  &  53.62       \\
        
        ${\text{JET}^\text{t}}$&70.39  &51.68   &59.72    &57.98&36.33&44.67&61.99&43.74&51.29 & 68.99 & 51.18 &  58.77       \\
        ${\text{JET}^\text{o}}$&62.26 & 56.84 &59.43&52.01&39.59&44.96&63.25&46.15&53.37 & 66.58  & 57.85 &  61.91\\
        \midrule
       
        ${\text{JET}^\text{t}_{\text{+}\text{BERT}}}$ &70.20    &53.02      &60.41&51.48&42.65&46.65&62.14&47.25  & 53.68  & 71.12& 57.20&63.41  \\
        
         ${\text{JET}^\text{o}_{\text{+}\text{BERT}}}$ &67.97  &60.32   &63.92      &58.47&43.67&50.00&58.35&51.43&54.67  & 64.77  & 61.29& 62.98  \\

         \midrule
         
       {${\text{Ours}_{\text{+}\text{BERT}}}$}  &73.96 & 67.87   &  70.78 &65.21&60.82&  62.94 & 64.86 &63.30 &  64.07 & \bf73.71  & \bf76.56& \bf 75.11 \\
        \bf {${\text{Ours}_{\text{+}\text{ALBERT}}}$}  &\bf77.32 & \bf75.52   &  \bf76.41 &\bf68.65&\bf61.22& \bf 64.72 & \bf68.36 & \bf66.81 & \bf 67.18 & 73.18  & 73.33&  73.25 \\
        \bottomrule
    \end{tabular}}
    \caption{Results on \emph{ASTE-Data-V1} test datasets. Baseline results are directly retrieved from \cite{xu-etal-2020-position}.}
    \label{aste_v1_results}
\end{table*}

\begin{table*}
\centering
\normalsize
\small
\begin{tabular}{lrrrrrrrr}
    \toprule
    \multirow{2}{*}{\normalsize{\textbf{Dataset}}} & \multicolumn{2}{c}{\normalsize{\textbf{\it14Rest}}}&\multicolumn{2}{c}{\normalsize{\textbf{{\it14Lap}}}}&\multicolumn{2}{c}{\normalsize{\textbf{\it15Rest}}}&\multicolumn{2}{c}{\normalsize{\textbf{\it16Rest}}}\\ 
     &\small{Sentences}&\small{Target}&\small{Sentences}&\small{Target}&\small{Sentences}&\small{Target}&\small{Sentences}&\small{Target}\\
	  
    \midrule
    ASTE-Data-V1-Train & 1,300 &2,145&920 & 1,265&593&923&842&1,289\\  
    ASTE-Data-V1-Valid  & 323&524 &228 & 337&148&238&210&316 \\  
    ASTE-Data-V1-Test& 496&862&339 & 490&318&455&320&465 \\
    \midrule
    ASTE-Data-V2-Train & 1,266 &2,338&906 & 1,460&605&1,013&857&1,394\\  
    ASTE-Data-V2-Valid  & 310&577 &219 & 346&148&249&210&339 \\  
    ASTE-Data-V2-Test& 492&994&328 & 543&322&485&326&514 \\
    \bottomrule
\end{tabular}
\caption{Statistics of the datasets used for the ASTE task.}  
\label{dataset_statistic_main}
\end{table*}

\begin{table}[h]
    \centering
    \small
    \begin{tabular}{lrrr}  
    \toprule  
    \textbf{Datasets} & \textbf{Sentence}  & \textbf{Aspect} & \textbf{Opinion} \\  
    \midrule  
    Restaurant14-Train  & 3,044 & 3,699 & 3,484 \\
    Restaurant14-Test   & 800   & 1,134 & 1,008 \\ 
    \midrule
    Laptop14-Train      & 3,048 & 2,373 & 2,504 \\ 
    Laptop14-Test       &   800 &   654 &   674 \\
    \midrule
    Restaurant15-Train  & 1,315 & 1,199 & 1,210 \\
    Restaurant15-Test   &   685 &   542 &   510 \\
    \bottomrule  
    \end{tabular} 
    \caption{Statistics of the datasets used for the AESC task.}
    \label{dataset_statistic_racl}
\end{table}

\subsection{Baselines}\label{baseline_intro}

Our model will compare to the following baselines on the ASTE task.

1) \textbf{RINANTE+} \cite{DBLP:conf/aaai/PengXBHLS20}. The model RINANTE is modified from that by \citet{ma-etal-2018-joint}. RINANTE+ is an LSTM-CRF model which first uses dependency relations of words to extract opinion and aspects with the sentiment. Then, all the candidate aspect-opinion pairs with position embedding are fed into the Bi-LSTM encoder to make a final classification.

2) \textbf{CMLA+} \cite{DBLP:conf/aaai/PengXBHLS20}. The model is adjusted from the one by \citet{DBLP:conf/aaai/WangPDX17}, which is an attention-based model, following the same two-stage processing with dependency relations as RINANTE+.

3) \textbf{Li-unified-R} \cite{DBLP:conf/aaai/PengXBHLS20}. Li-unified-R utilizes a modulated multi-layer LSTM encoder by \citet{li-lu-2019-learning}, and adopts the same aspect-opinion pair classification as RINANTE+.

4) \textbf{Peng et al.} \cite{DBLP:conf/aaai/PengXBHLS20}. This model adopts GCN to capture dependency information, and at the second stage, uses the same strategy of RINANTE+ to fulfill triplet extraction.
  
5) \textbf{OTE-MTL} \cite{zhang-etal-2020-multi-task}. A multi-task learning approach that incorporates word dependency parsing boosts the performance of triplet extraction.
  
6) \textbf{JET} \cite{xu-etal-2020-position}. This model jointly extracts all the subtasks through a unified sequence labeling method. ${\text{JET}^\text{t}}$ and ${\text{JET}^\text{o}}$ denote two different tagging forms.

7) \textbf{GTS} \cite{wu-etal-2020-grid}. A sequence tagging model leverages the property element upper triangular matrix to model the extraction of aspect and opinion terms.
 
8) \textbf{Huang et al.} \cite{huang2021first}. The latest sequence labeling model which utilizes the restricted attention field mechanism and represents word-word perceivable pairs for the final classification.

For the AESC task, our model will compare to the following baselines:
 
1) \textbf{SPAN-BERT} \cite{hu-etal-2019-open}. It is a BERT-based model which utilizes span representation to perform the AESC task. 
 
2) \textbf{IMN-BERT} \cite{hu-etal-2019-open}. It is a multi task learning model modified by \citet{he-etal-2019-interactive} and utilizes BERT as encoder to perform aspect term extraction and sentiment classification.

3) \textbf{RACL-BERT} \cite{chen-qian-2020-relation}. It is a multi-layer multi-task learning model with mutual information propagation to boost the performance of the AESC task.

4) \textbf{Mao et al.} \cite{mao2021joint}.
It is a dual-MRC architecture model to detect the AT/OT and corresponding sentiment polarity by means of a two-round query answering approach. 

\subsection{Results on ASTE-Data-V1 for ASTE}

Results on the \emph{ASTE-Data-V1} datasets also show the effectiveness of our model. But there is an interesting phenomenon that on the {\it16Rest} test set, the result of ALBERT-based model is lower than that of BERT-based model. It may be due to the inconsistent domain between the test set and the pre-trained language model.

\subsection{Data Statistics}

Table \ref{dataset_statistic_main} and Table \ref{dataset_statistic_racl} show the statistics of the datasets we used.

\end{document}